\documentclass[letterpaper, 10 pt, conference]{ieeeconf}  % Comment this line out
\IEEEoverridecommandlockouts                              % This command is only
\usepackage{graphics} % for pdf, bitmapped graphics files
\usepackage{epsfig} % for postscript graphics files
\usepackage{mathptmx} % assumes new font selection scheme installed
\usepackage{times} % assumes new font selection scheme installed
\usepackage{amsmath} % assumes amsmath package installed
\usepackage{amssymb}  % assumes amsmath package installed
\usepackage{mathtools}
\usepackage[noadjust]{cite}
\usepackage{dblfloatfix}
\usepackage{color}
\usepackage{booktabs,tabularx}
\usepackage{optidef}
\usepackage[font=small,skip=3pt]{caption}

\renewcommand{\thefigure}{\arabic{figure}}
\title{\LARGE \bf
Fabric Soft Poly-Limbs for Physical Assistance of Daily Living Tasks  
}

\author{Pham H. Nguyen, \textit{Student Member, IEEE}, Imran I. B. Mohd, Curtis Sparks, Francisco L. Arellano,\\ Wenlong Zhang$^{*}$, \textit{Member, IEEE} and Panagiotis Polygerinos, \textit{Member, IEEE}% <-this % stops a space
\thanks{$*$ Address all correspondence to this author.}% <-this % stops a space
\thanks{Pham H. Nguyen, Curtis Sparks, Wenlong Zhang and Panagiotis Polygerinos are with the Polytechnic School, Ira A. Fulton Schools of Engineering, Arizona State University, Mesa, AZ 85212, USA. 
         {\tt\small nhpham2@asu.edu; cmspark1@asu.edu; polygerinos@asu.edu}}%
\thanks{Imran I. B. Mohd The School for Engineering of Matter, Transport and Energy, Ira A. Fulton Schools of Engineering, Arizona State University, Tempe, AZ 85281, USA.
        {\tt\small imohd@asu.edu}}%
\thanks{Francisco L. Arellano The School of Biological Health Systems Engineering, Ira A. Fulton Schools of Engineering, Arizona State University, Tempe, AZ 85281, USA.
        {\tt\small flopezar@asu.edu}}%
}

\begin{document}

\maketitle
\thispagestyle{empty}
\pagestyle{empty}

%%%%%%%%%%%%%%%%%%%%%%%%%%%%%%%%%%%%%%%%%%%%%%%%%%%%%%%%%%%%%%%%%%%%%%%%%%%%%%%%
\begin{abstract}

This paper presents the design and development of a highly articulated, continuum, wearable, fabric-based Soft Poly-Limb (fSPL). This fabric soft arm acts as an additional limb that provides the wearer with mobile manipulation assistance through the use of soft actuators made with high-strength inflatable fabrics. In this work, a set of systematic design rules is presented for the creation of highly compliant soft robotic limbs through an understanding of the fabric based component’s behavior as a function of input pressure. These design rules are generated by investigating a range of parameters through computational finite-element method (FEM) models focusing on the fSPL's articulation capabilities and payload capacity in 3D space. The theoretical motion and payload outputs of the fSPL and its components are experimentally validated as well as additional evaluations verify its capability to safely carry loads 10.1x its body weight, by wrapping around the object. Finally, we demonstrate how the fully collapsible fSPL can comfortably be stored in a soft-waist belt and interact with the wearer through spatial mobility and preliminary pick-and-place control experiments.

 %This soft arm acts as an additional limb that provides compliant and safe mobile manipulation assistance to users with upper extremity impairments.
 %of the f3CBAs 
% * <polygerinos@asu.edu> 2018-09-14T16:49:09.966Z:
% 
% > This soft arm acts as an additional limb that provides compliant mobile manipulation assistance through the use of soft actuators made with high-strength fabrics.
% how about this instead? since we are not really touching the impairment aspect of this project here.
% 
% ^.
 %of the FEM models 

%investigated through experimental validations 
\end{abstract}

%%%%%%%%%%%%%%%%%%%%%%%%%%%%%%%%%%%%%%%%%%%%%%%%%%%%%%%%%%%%%%%%%%%%%%%%%%%%%%%%
\section{INTRODUCTION}
%\vspace{-0.75em}
% Cervical Spondylotic Myelopathy (CSM) is a degenerative condition that is caused by general wear and tear or injury to the cervical region of the spinal cord, affecting limb functions and limiting independence \cite{lubelski2016}. As a result, individuals with CSM syndrome typically experience difficulty performing motor tasks, such as activities of daily living (ADLs). This population could benefit from a soft wearable collaborative robotic device, such as the fabric-based, Soft-Poly Limb (fSPL) presented in this work.

Individuals with upper-limb impairments typically experience difficulty performing motor tasks, such as activities of daily living (ADLs). These populations could benefit from a soft wearable collaborative robotic device, such as the fabric-based, Soft-Poly Limb (fSPL) presented in this work.

\begin{figure}[t!]
\centering
\includegraphics[width=0.35\textwidth]{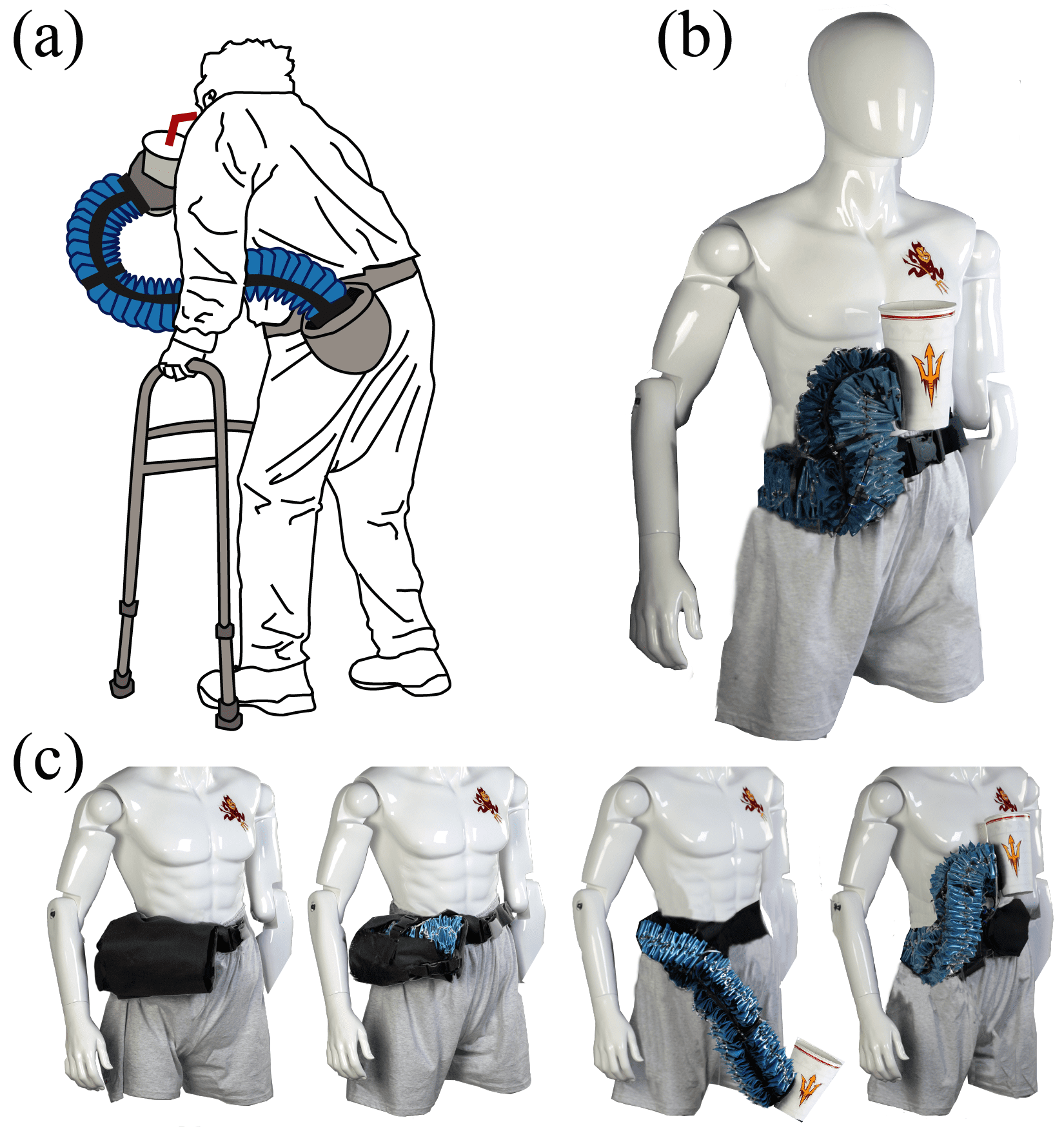}
\setlength{\belowcaptionskip}{-26pt}
\caption{ (a) Illustrated concept of the fSPL. (b) The prototype fSPL. (c) Left-to-right: the limb is collapsed and stored in a pouch and then deployed to assist with tasks, such as picking up a cup.}
\label{fig:fig1}
% \vspace{-1.5em}
\end{figure}

% (a) Illustrated concept of the fabric-based Soft Poly Limb (fSPL). (b) The prototype fSPL mounted on a lightweight waist-belt system is a soft, lightweight, wearable robot that is designed to assist users with activities of daily living (ADL). (c) Left-to-right: the limb is collapsed and stored in a pouch and then deployed to assist with tasks, such as picking up a cup.

Wearable robotics manipulators are devices that can be worn to provide an additional appendage to assist and support the user to perform tasks. These devices do not kinematically have to match the human body's anatomy, like exoskeletons \cite{gopura2011}, or prosthetic devices \cite{bogue2009}. Recent examples of different wearable collaborative robotic devices include robotic legs \cite{kurek2017,parietti2015}, arms \cite{parietti2014,saraiji2018,vatsal2017}, fingers \cite{hussain2017b,wu2015,tiziani2017}, ranging from industrial \cite{parietti2014} to medical applications \cite{tiziani2017}. 

% The recent rise of wearable robotic manipulators have brought up questions of user controllability. The possibility of controlling an extra limb has been previously investigated and it is known that the human central nervous system (CNS) is found to be capable of learning to control additional limbs \cite{guterstam2011,tsakiris2010}. The controllability of robotic limbs has been explored in preliminary research, utilizing biological signals from the torso muscles \cite{parietti2017}, foot \cite{sasaki2017}, elbow \cite{wu2015}, and the forehead frontalis muscles \cite{salvietti2017}.

Wearable manipulators generally face limitations that originate from their rigid designs, like their weight, bulk, and the interaction between rigid elements and the human body \cite{delAma2012}. Soft robotics has emerged as one of the solutions to tackle these aforementioned challenges. It has introduced a variety of soft continuum manipulators subdivided based on actuators that constitute them, including cable-driven \cite{mcMahan2005,calisti2011}, pneumatic artificial muscles (PAMs) \cite{walker2005,godage2016,yasmin2017,giannaccini2017}, elastomeric \cite{cianchetti2013,marchese2015,robertson2017,gong2018}, origami \cite{santoso2017}, and inflatable-fabric \cite{sanan2013,hawkes2017,ohta2017,best2016,liang2018,takeichi2017,kim2018} based manipulators. 
% * <polygerinos@asu.edu> 2018-09-09T23:28:36.896Z:
% 
% > The recent influx of soft robots have led to the introduction a variety of soft continuum manipulators made of a variety of soft actuators. The categories of soft continuum manipulators is subdivided based on the actuators that constitute them, including cable-driven \cite{mcMahan2005,calisti2011}, pneumatic artificial muscles (PAMs) 
% it would have been great to have your review paper cited here....if they accept we can add in january 
%Berm:Yes PANOS!!
% 
% ^.

The fusion of soft robotics and wearable manipulators has created a new category of robotics which we call, Soft Poly-Limbs (SPLs) \cite{nguyen2018}. Limited research has started to emerge in this category, Tiziani et al. \cite{tiziani2017} has introduced a soft poly-finger device, Liang et al.~\cite{liang2018} have suggested a two-DOF fabric-based soft poly-arm device that would eventually be integrated as a wearable robot, but have not evaluated the device for this use case yet. Finally, we recently introduced an elastomeric SPL device \cite{nguyen2018} capable of wrapping around objects to lift approximately 2.35x its own weight. 

In this paper, we have designed a novel, highly maneuverable, lightweight SPL capable of assisting users made entirely of fabric (Fig.~\ref{fig:fig1}) capable of lifting up to $1.5kg$, while counteracting gravity in an outstretched position. We further investigate the mechanical behavior of the fabric-based SPL (fSPL) and its components by using a set of design parameters to optimize the payload capacity of the fSPL by using new computational finite element method (FEM) models that are validated experimentally. We also assess the large degrees of freedom of the fSPL through a workspace experiment and its ability to carry load, up to $10.1x$ its body weight, by wrapping around an object. Finally, through a series of experiments we showcase the potential of the fSPL, capable of compressing to $1/2$ its entire length, to work safely around the wearer and assist with ADLs.

\label{sec:func_req}
\begin{table}[t!]
\caption{Requirements for a fabric-based Soft-Poly Limb (fSPL)} 
\label{tab:spec_table}
	\begin{tabularx}{0.48\textwidth}{*2l}    \toprule\toprule
	\textbf{\emph{Characteristics}} & \textbf{\emph{Requirements}} \\\midrule
	Weight of Device    & Approx. 1.0$kg$  \\ 
	Profile of fSPL & Less than 100$mm$ diameter \\ 
	Payload (fully extended)    & Approx. 1$kg$  \\ 
	Payload (whole-body grasp) & 2x the weight of device \\ 
	Max. Reach Length    & Length of male arm (0.59$m$)  \\ 
	Min. Retraction Length & Half of fully stretched SPL \\ 
	Safety    & Easy to don-and-doff \\ 
	 &Does not interfere with\\
     &biological limb's movement\\
	Degrees-of-Freedom (DOF) & Infinite DOF (continuum) \\ 
	Degrees of Curvature of Each Segment & 180$^{\circ}$ \\\bottomrule
	 \hline
	\end{tabularx}
    \vspace{-5mm}
\end{table}

\begin{figure}[!hb]
\vspace*{-0.5cm}
\centering
\includegraphics[width=0.40\textwidth]{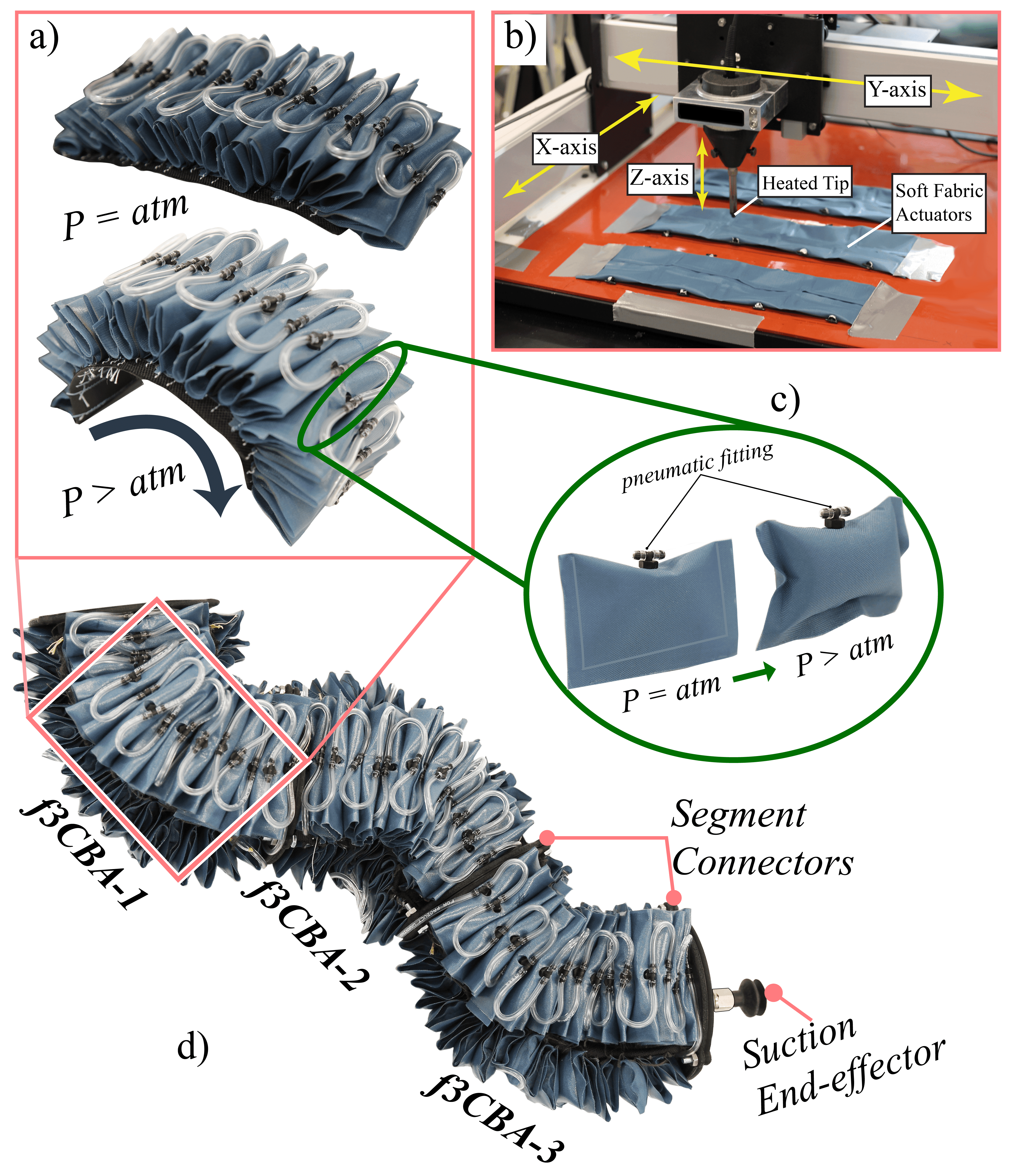}
\caption{a) The actuator array when inflated. b) The CNC Process to fabricate fabric actuators. c) Each singular fabric actuator. d) The entire fSPL composed of 3 segments of f3BAs.}
\label{fig:fabrication}
% \vspace{-10mm}
\end{figure}

\section{Design and Fabrication of fSPL}

\subsection{Functional Requirements}
\label{sec:func_req}
In order to provide long-term options to users who have lost/reduced function in their hands and arms we have laid out a soft robotic architecture with functional requirements, as summarized in Table~\ref{tab:spec_table}. We intend to provide the wearer with a full-length SPL that interacts safely with the user and the environment. Because of the inflatable fabric materials we intend to utilize for its structure and motion, we anticipate a lightweight (approximately 1$kg$) and slim body that can be collapsed, several times its original length, and stowed away in a pouch attached to a waist belt without adding bulkiness or impeding the user's motion. In general, the hyperredundant SPL should be a highly maneuverable manipulator exceeding the biological arm capabilities by offering infinite number of kinematic DOFs. This can be achieved by utilizing soft continuum designs, which could also offer configurable and compliant motions. Considering manipulation of most daily living objects, the system should also be able to carry up to approximately 1$kg$ of payload at its end-effector while being fully-extended and positioned parallel to the ground. Finally, because of its soft and continuum nature, the arm would be able to achieve whole-body grasping and manipulate even higher loads by wrapping around objects and distributing the loads. 
% * <polygerinos@asu.edu> 2018-09-11T20:03:46.580Z:
% 
% > with 9-DOFs
% still i am not sure about this number...currently we do not provide explanation as to why 9dof is ideal..consider deleting the number...
%  I changed it to at least 9-DOFs

% Berm:If you think it is still necessary to to provide it then we can delete it. Because the other arm papers specify their DOFs and they are all less than what we are dealing with, so I thought it is important to mention.

% "In general, an SPL should be a highly maneuverable continuum manipulator with at least 9-DOFs, as its continuum soft structure can provide an infinite number of DOFs, while being compliant and configurable "
% ^.

\subsection{Material Selection}
\label{sec:mat_select}

The applicability and functionality of fabric bending actuator arrays have been highlighted in our previous work~\cite{carly2018}, where thin thermoplastic polyurethane (TPU) actuators encased in nylon fabric casings are able to withstand pressures higher than 0.3$MPa$. Instead, in this work a variety of heat-sealable TPU-coated nylon fabric materials are explored to develop soft actuators that are robust to external environmental interactions, allow for a single-step fabrication process to save manufacturing time, and are capable of withstanding even higher pressures (up to 0.53$MPa$), thus achieving higher bending and torque outputs.
% * <polygerinos@asu.edu> 2018-09-11T23:36:58.827Z:
% 
% > work \cite{thalman2018}
% there is something wrong with this citation..please fix
% Berm: I have fixed it
% ^.

\begin{table}[t!]
\caption{Material Properties of TPU Coated Nylon Fabrics} 
\label{tab:materialproperty_table}
	\begin{tabularx}{0.48\textwidth}{l|c|c|c}   \toprule\toprule
%     \centering
    \small
    \setlength\tabcolsep{11pt}
	\textbf{\emph{Material}} & \textbf{\emph{Density}} & \textbf{\emph{Seal Strength}} & \textbf{\emph{Burst }} \\[-1pt]
                             & \textbf{\emph{(kg/m$^3$)}}                             & \textbf{\emph{(N)}} & \textbf{\emph{Strength}}\\
                             &                              &                            & \textbf{\emph{(MPa)}}\\\midrule
	Rockywoods 200D 	 &	840.00 		&	168.35 	&0.53$\pm$0.04 \\
    Outdoor Oxford 200D  &	757.58 		&  	183.68 	&0.48$\pm$0.03 \\
    Ripstop 200D 		 &	758.62 		& 	192.19 	&0.40$\pm$0.017	\\
    Seatle Diamond 200D  &	892.86 		& 	164.75	&0.36$\pm$0.026 \\
    DIY Packraft 400D 	 & 	982.46		& 	155.32 	&0.20$\pm$0.026\\
    DIY Packraft 1000D	 & 	1000.00 	& 	236.17 	&0.11$\pm$0.02\\
    Taffeta 70D 		 &	700.00		& 	150.59  &0.048$\pm$0.008\\\bottomrule 
    \hline
	\end{tabularx}
    \vspace{-6mm}
\end{table}

A set of TPU-coated nylon fabrics with denier values ranging from 70D to a 1000D are characterized against seal strength peel and burst tests, as seen in Table \ref{tab:materialproperty_table}. It is noted that the dernier number indicates the fiber thickness of the filament used for their fabrication.

% We perform a seal strength peel test, following the ASTM F88/F88M-15 protocol. From these tests we select the top four materials demonstrating the highest seal strength and lowest densities and further test them for the maximum pressure they could withstand when inflated (burst test), using the ASTM F2054 protocol. Based on the results, the Rockywoods 200D heat-sealable fabric (6607, Rockywoods Fabric, Loveland, CO) is chosen for its highest burst strength of 0.53$MPa$. 

We perform a seal strength peel test, with the ASTM F88/F88M-15 protocol. We then further test them for the maximum pressure they could withstand until failure (burst test), using the ASTM F2054 protocol. Based on the results, the Rockywoods 200D heat-sealable fabric (6607, Rockywoods Fabric, Loveland, CO) is chosen for its highest burst strength of 0.53$MPa$. 
%     Finally, the elastic Young's modulus for the selected material is determined using the ASTM D882 protocol and found to be 489.9$MPa$.

% * <polygerinos@asu.edu> 2018-09-09T23:52:24.093Z:
% 
% > 1$kg$ of payload at its end-effector position when fully-extended and parallel to the ground and carry up to twice i
% you do not explain why 1kg and why twice the load...please add and break into smaller sentences 
% 
% ^.
% * <polygerinos@asu.edu> 2018-09-09T23:48:19.888Z:
% 
% > with 9-DOFs
% the 9dofs is highly arbitrary and you do not provide any good explanation.
% plus still you havent decided if continuum arms can claim degrees of freedom like rigid arms do. still i think here we have infinite/multiple dofs  that offer high actuation redundancy.
% DECIDED and infinite DOFs
% ^.

\subsection{Design, Fabrication and Integration}
To meet the functional requirements, the length of the fSPL is designed to match the length of an average sized adult male arm, approximately 0.59$m$ \cite{plagenhoef1983}, from the tip of the shoulder to the center of the wrist. The fSPL is subdivided into three active segments called the fabric 3 bending actuators (f3BAs), as seen in Fig. \ref{fig:fabrication}d. Each segment further consists of three bending actuators arrays (shown in Fig. \ref{fig:fabrication}a arranged and sewn into an equilateral triangle prism to create the f3BA segment. %, using a sewing machine (Memory Craft 6500P, Janome, Hachioji, Tokyo). 
Each bending actuator array is comprised of multiple high-strength, heat-sealed, TPU-coated nylon (200D) fabric actuators (shown in Fig. \ref{fig:fabrication}c, previously selected in Section~\ref{sec:mat_select}. 
% * <polygerinos@asu.edu> 2018-09-11T23:42:18.160Z:
% 
% > , approximately 0.59$m$, from the tip of the shoulder to the center of the wrist. 
% maybe add citation?
% Berm: Added citation
% ^.

To develop the fabric actuators, the selected material is cut into a rectangular shape using a laser cutter (Glowforge Pro, Glowforge, Seattle, WA) and the pneumatic fittings are attached (5463K361, McMaster-Carr, Elmhurst, IL). A sealant is also added around the fittings to prevent air leakage (Seam Grip, Gear Aid, Bellingham, WA). The material is arranged on a customized CNC router (Shapeoko 3, Carbide Motion, Torrance, CA) with a soldering iron tip set at 230$^{\circ}$C and traced to seal the fabric actuators, as seen in Fig. \ref{fig:fabrication}b. This apparatus enables rapid heat-sealing of fabric soft actuator patterns accurately and repeatedly, up to 25 individual actuators at a time for this limb design. 

Fabric pockets to slot the individual actuators are evenly sewn into a parallel array configuration onto a strain-limited fabric layer. The individual actuators as depicted in Fig. \ref{fig:fabrication}c, are inserted into the pockets, creating actuator arrays. When the actuators inflate, their interaction causes the actuator array to bend or even curl completely, as shown in Fig.~\ref{fig:fabrication}a. 
% * <polygerinos@asu.edu> 2018-09-11T23:51:24.373Z:
% 
% > This approach allows for ease of replacement of the actuators during maintenance. 
% you  could omit this sentence and instead briefly speak about the principle of operation of this array. how is this design producing the bending motions?
% Berm: Have added the bending motion part

% Berm:I have added a line that I cut before to replace it:
% " When the actuators on the actuator array inflate, their interaction causes the actuator array to bend or even curl completely, as shown in Fig. \ref{fig:fabrication}a. "

% ^.

Modular 3D-printed connectors are sewn into the distal and proximal ends of each f3BA and used to connect all the segments, creating the complete fSPL (Fig. \ref{fig:fabrication}d). The modularity of the connector pieces allow for a variety of end-effectors to be mounted at the distal end of the fSPL, such as a suction cup (5427A636, McMaster-Carr, Elmhurst, IL). A zippered fabric pouch is designed to store the fSPL by collapsing its body to less than half its full length. To interface the fSPL with the human body, the fabric pouch is sewn onto a modified technical belt (S\&F Deluxe Technical Belt, Lowepro, Petaluma, CA) as shown in Fig. \ref{fig:fig1}c. This 0.5$kg$ waist belt, designed to be don-doff friendly, ensures that loads carried by the fSPL can be evenly distributed along the waist of the wearer without creating pressure points.

\begin{figure}[b!]
\centering
\includegraphics[width=0.41\textwidth]{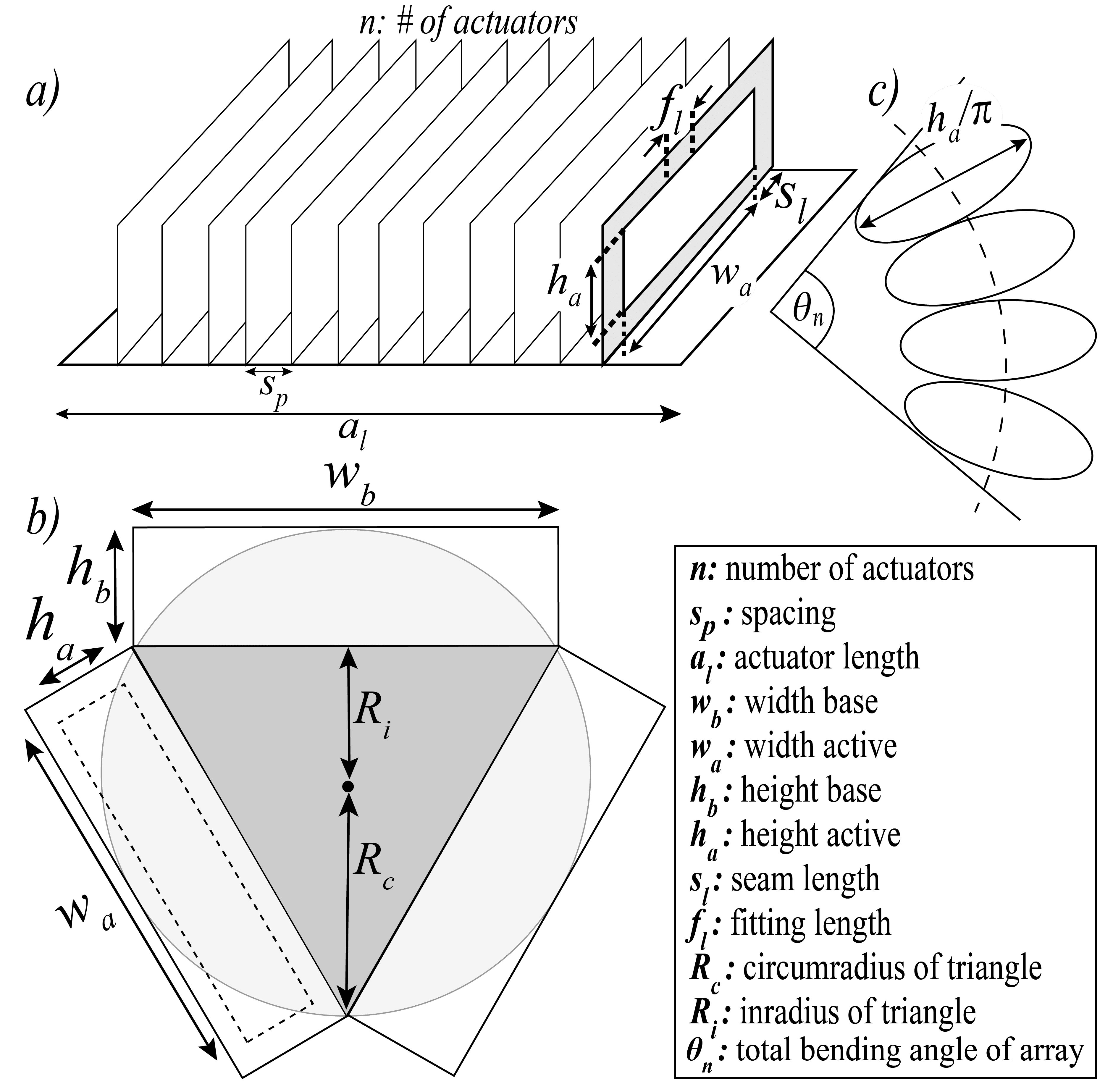}
\caption{a) Isometric view of actuator array. b) Bottom view of f3BA. c) Side view of bending actuator array.}
\label{fig:geom_param}
% \vspace{-2.2em}
\end{figure}

\section{fem-based Optimization of fSPL}
An FEM modeling approach is used to predict the bending performance and tip payload force of the fSPL and its components throughout this work. It is also used to identify the geometrical parameters, as seen in Fig.~\ref{fig:geom_param}, that maximize the tip force of the actuator array design. In this work, we introduce the use of ABAQUS/Explicit (Simulia, Dassault Systemes) and shell elements to model inflatable thin plastic films. Because of the large deformations and short dynamic response times observed amongst the fabric actuators, convergence of computational solutions can be only obtained through an explicit (dynamic) simulation environment.

The materials properties used in our simulations include the linear elastic modulus, determined with ASTM D882, of the TPU-coated nylon selected in Section~\ref{sec:mat_select} (Young's modulus of E = 498MPa and Poisson’s ratio of v = 0.35), the properties for the inextensible fabric layer used to hold the actuators evenly-spaced in an array  (E = 305MPa, v = 0.35), and the properties used for the PLA connector caps (E = 3600, v = 0.3). All the components of the actuators are modeled using shell explicit quadratic tetrahedral elements (C3D10M).  

Although the explicit solution is a true dynamic procedure, it has the valuable capability of solving quasi-static contact problems as well. In our case, it would be the actuator arrays inflating to contact the force plate to measure the tip payload force. Because it is impractical to simulate the event in a natural time scale as it would be computationally demanding, the explicit solution would need to be accelerated, while still maintaining its dynamic equilibrium. The first requirement is to ensure loading is as smooth as possible, so not to introduce noise. This is achieved by controlling the loading rate of the analysis at 1\% of the speed of the stress wave of the material. To verify if quasi-static conditions are achieved in the simulation, the requirement is to inspect that the total kinetic energy of the deforming material does not exceed 5\% of its total internal energy (IE) throughout the simulation.

%  or setting the time period to at least 10x the natural period of the model  Thus, ALLKE (kinetic energy) must be just 5\% of the ALLIE (internal energy).

\subsection{Geometrical Parameters of Fabric Actuators }
% * <polygerinos@asu.edu> 2018-09-10T00:18:39.923Z:
% 
% > Bladder 
% be consistent with the names you use...this is the first you call something a bladder. please revise
% Berm: I have changed this.
% ^.
To study the payload and bending capabilities of the fSPL prior to fabrication, a set of geometrical parameters, as highlighted in Fig. \ref{fig:geom_param}, are studied. The three main geometrical parameters studied and optimized are the active (i.e. without the seam length) width and height of the actuators ($w_{a}$ and $h_{a}$), and the number of actuators ($n$) based on their major contribution to the bending and payload capacity of the fSPL.

The active height ($h_{a}$) contributes to the bending capability, while the active width ($w_{a}$) contributes to the stability of the actuator array of length, $a_{l}$. If $w_{a}$ is less than $h_{a}$ then the contact surface between consecutive actuators will be smaller and therefore the actuators will lose friction resulting in torsional motions and decreased force distribution.

Therefore, $w_a$ is set to be greater than $h_a$ by a ratio $r$ that is larger or equal to at least $1.0$. The active width and height are also constrained by the diameter of the physical tube fitting $f_l$ that will be used, which is 6.2$mm$, and the heat seam width $s_l$ created by the heat sealer, which is 5$mm$. The spacing between the actuators ($s_{p}$) is limited to over $7.5mm$ because of sewing manufacturing limitations. Both the active width and height of the actuator are also constrained by $R_{c}$, which defines the cross-sectional radius of the fSPL, constrained to 50$mm$. The angle of the inflated  actuator array  $\theta_{n}$, is designed to achieve at least 180$^{\circ}$ when fully inflated. The aforementioned  physical constraints are summarized below:
\begin{align*}
R_c &= 50mm,    &  s_l &\geq 5mm,      &  \theta_n&=180^\circ,\\
f_l&=6.23mm,         &  a_l&=160mm,   &  s_p&\geq7.5mm.
\label{eq:constraints}
\end{align*}

Using the number of actuators ($n$) and the spacing ($s_{p}$), the active height ($h_{a}$) can be calculated as follows:
% * <polygerinos@asu.edu> 2018-09-10T00:27:02.256Z:
% 
% > bladders 
% again the word bladders..
% Berm: NO MORE BLADDERS
% ^.
\begin{align} 
w_a &= r\cdot h_a, \\
s_p &=  \frac{(a_{l} - 2\cdot(a_l/n))}{n}, \\ 
h_a &=  \frac{s_p \cdot \pi}{2\cdot(1-sin(\theta_n/(2\cdot n))} \label{eq:wa_sp_ha}.
\end{align}

From the circumradius ($R_c$) and inradius ($R_i$) of Fig. \ref{fig:geom_param}b, we can calculate the active width ($w_a$) as:
\begin{align} 
w_a &\leq  R_c/(\frac{\sqrt{3}}{6} + \frac{1}{r}) - 2\cdot s_l -f_l \label{eq:wa_ha}.
\end{align}

\begin{figure}[t!]
% \vspace*{-0.65cm}
\centering
\includegraphics[width=0.41\textwidth]{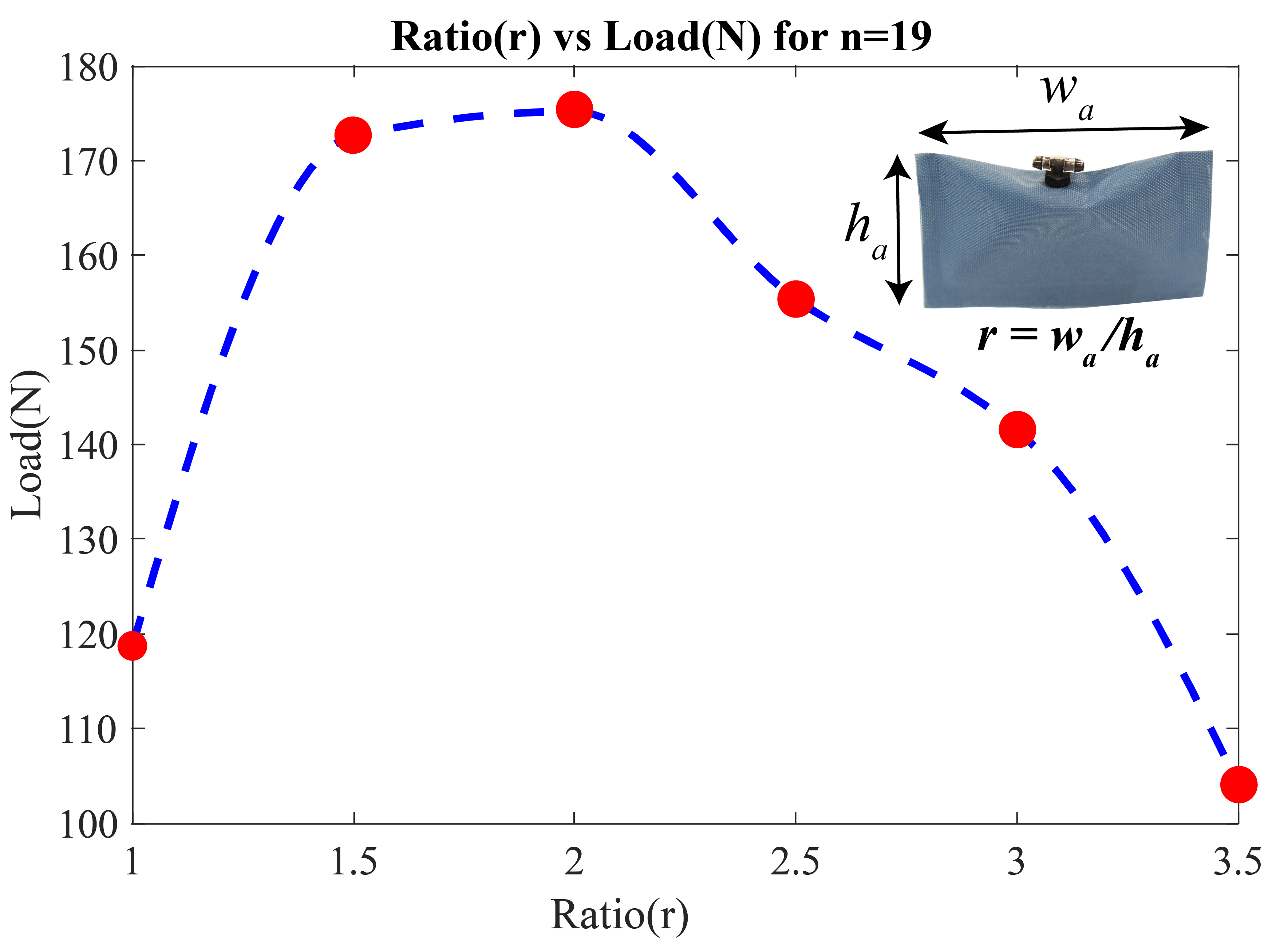}
\caption{FEM analysis of number of actuators and ratio ($r = w_a/h_a$) with regards to the force (N) generated by actuator array.}
\label{fig:fem_opti}
\vspace{-1.5em}
\end{figure}

\begin{figure}[b!]
\vspace*{-0.5cm}
\centering
\includegraphics[width=0.41\textwidth]{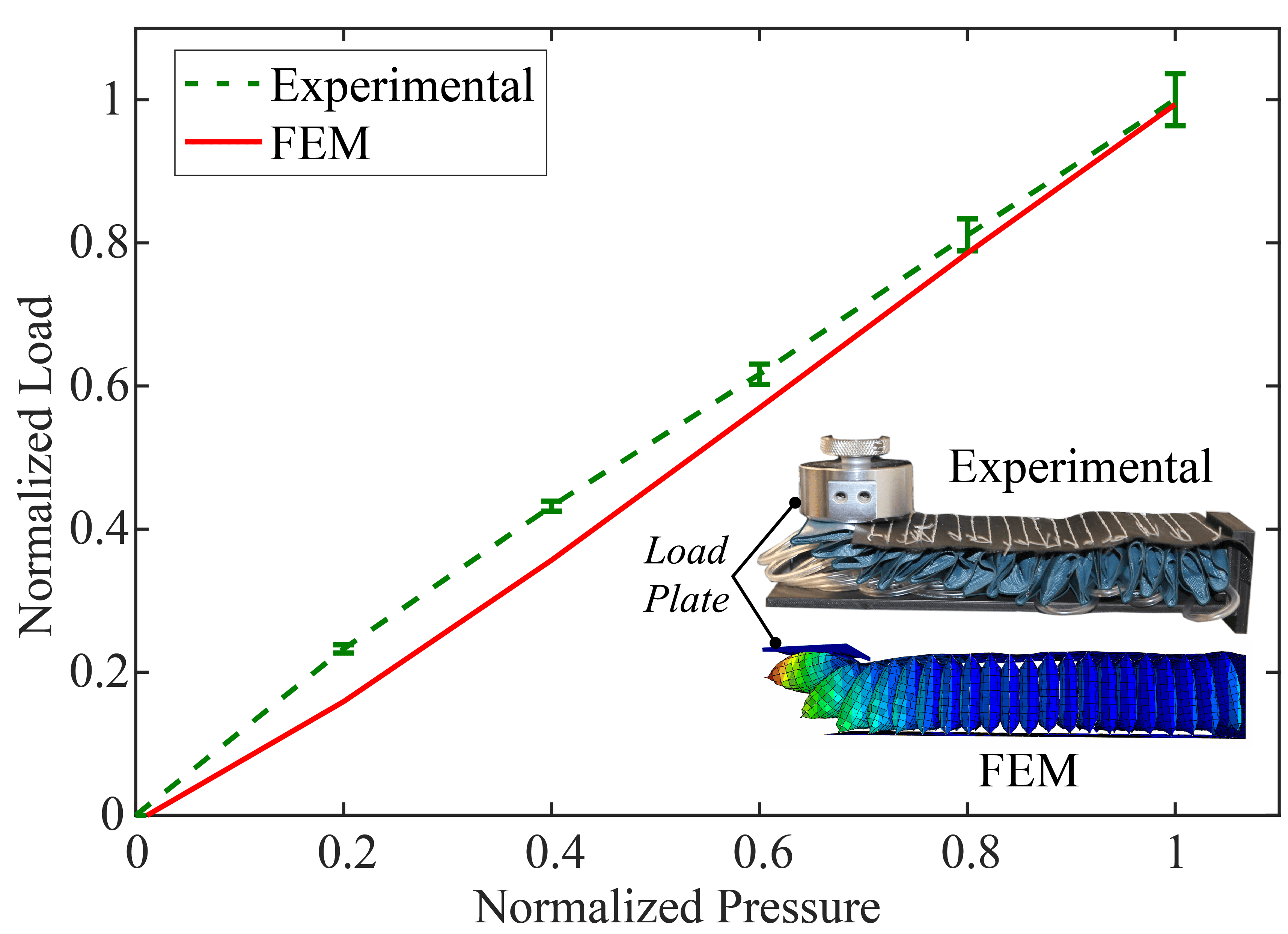}
\caption{Payload capability of a single actuator array compared between FEM and experimentally validated, repeated 3 times}
% * <polygerinos@asu.edu> 2018-09-14T18:11:58.104Z:
% 
% > Payload capability of a Single Bladder Array compared between FEM and Experimentally Validated}
% bladder again!!!!
% It is gone
% ^.
\label{fig:single_fem_real}
% \vspace{-1.5em}
\end{figure}

\begin{figure}[b!]
\vspace*{-0.5cm}
\centering
\includegraphics[width=0.41\textwidth]{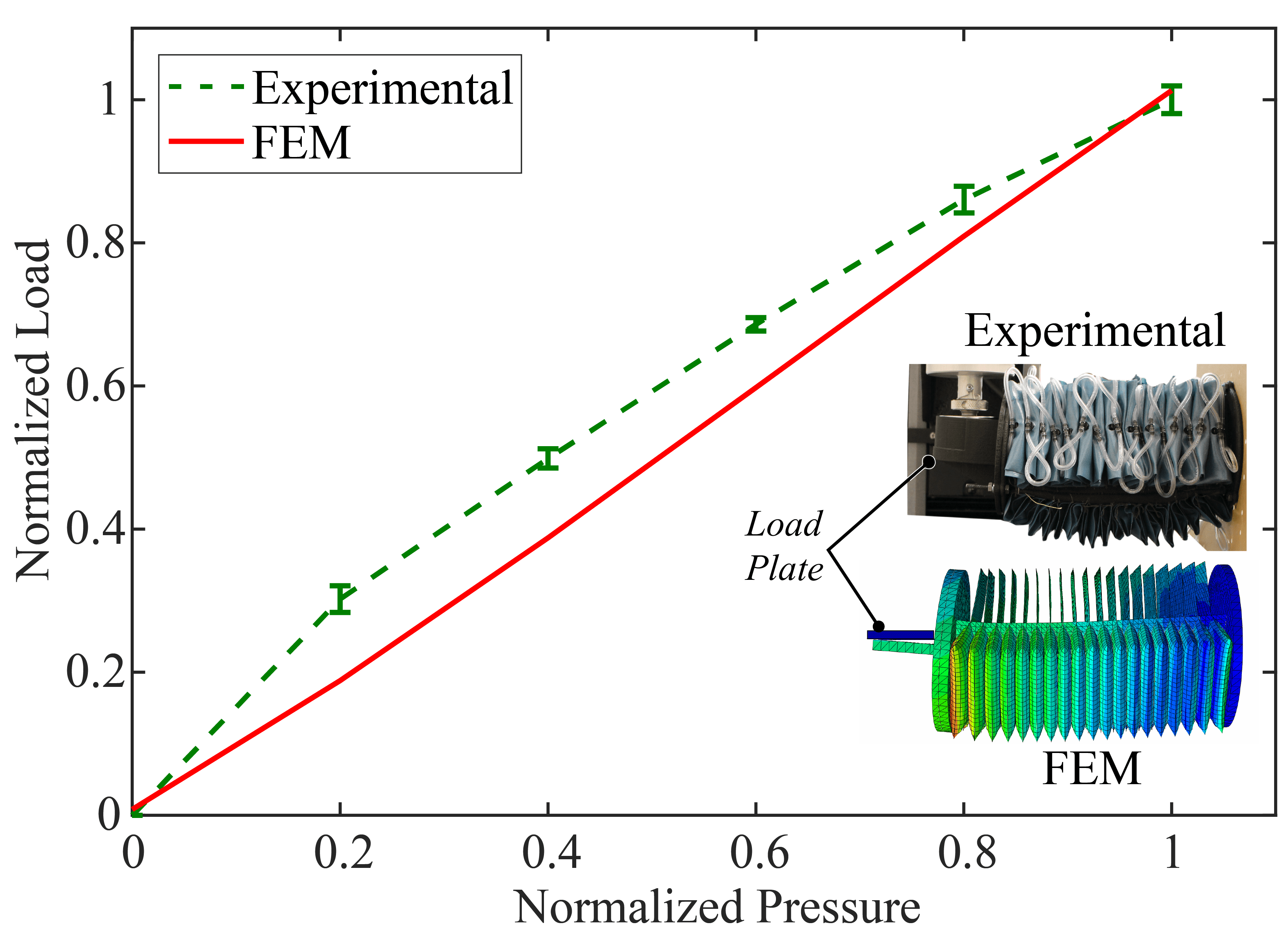}
\caption{Payload capability of a f3BA unit compared between FEM and experimentally validation, repeated 3 times.}
% * <polygerinos@asu.edu> 2018-09-14T18:24:11.913Z:
% 
% > \caption{Payload capability of a f3CBA unit compared between FEM and Experimentally Validated}
% > \label{fig:f3CAs_load_fem_real}dont forget to mention the number of times the test was perfromed
%Berm: Have added in the next
% 
% ^.
\label{fig:f3CAs_load_fem_real}
% \vspace{-1.5em}
\end{figure}

\subsection{FEM Optimization of Geometrical Parameters}

% * <polygerinos@asu.edu> 2018-09-10T01:20:49.833Z:
% 
% > % \label{fig:single_fem_real}
% > % \vspace{-1.5em}
% 
% all good but you have forgotten to add a critical explanatory paragraph that speaks of how you made the FEM model....explicit, shells, element types...interactions...etc
% Berm: I have added
% ^.

To find the geometrical parameters required to maximize the tip force of the bending actuator array, we use FEM simulations with varying parameters, the ratio ($r$) and the number of actuators ($n$). The number of actuators ($n$) is varied based on the minimum and maximum $s_p$. The maximum $n$ is determined from (2) by the minimum $s_p \geq 7.5mm$, which demonstrates that $n = 19$ is the optimum number.  The minimum $n$ is determined by the number of actuators required to generate significant tip force from the actuator array as $n = 8$. Thus from our study, more actuators in the actuator array, contributes to a larger tip force. By varying the ratio ($r$) between $w_a$ and $h_a$, from $1.0$ to $3.5$, we notice that the tip force increases with the increasing ratio until a ratio of $2.0$ is reached. After which the force decreases because the radius of the inflated actuator becomes smaller than the spacing ($s_p$) between the actuators. Therefore, the configuration that produces the highest tip force ($n = 19$) and is less likely to to present actuator slippage is selected, with ratio ($r =2.0$), as seen in Fig.~\ref{fig:fem_opti}. 

\begin{table}[t!]
\caption{UTM and Free-space Bending Payload} 
\label{tab:payload_table}
	\begin{tabularx}{0.48\textwidth}{c|c| c|c c}   \toprule\toprule
    \centering
    \small
    \setlength\tabcolsep{11pt}
	\textbf{\emph{Component}} & \multicolumn{2}{c|}{\textbf{\emph{UTM Payload }}} & \textbf{\emph{Free-space Loading}} \\[-1pt]
                             &\multicolumn{2}{c|}{\textbf{\emph{(N) }}}  	&\textbf{\emph{Payload (N or kg)}}	\\\midrule
                            
					&	\textbf{Exp.}	&	\textbf{FEM}	&	\\\cline{2-3}\rule{0pt}{2.8ex}

    Actuator Array 	&		174.73$\pm$6.37 	&177.2 	 & -- \\
    f3BA			&		53.73$\pm$1.04		&56.9	 & 58.8 or 6.0\\
    fSPL			& 		14.91$\pm$0.93	&20.0	 & 14.7 or 1.5\\\bottomrule 
    \hline
	\end{tabularx}
    \vspace{-1.5em}
\end{table}
\section{Testing and Evaluation of fSPl}
For the fSPL to be effective during everyday use, it is important to carry the desired payload and maneuver it effectively within the 3D workspace of the wearer. Therefore, to assess the capability of the fSPL, we explore experimental validations of the FEM models, various payload experiments, the workspace of the fSPL, and finally users' ability to perform pick-and-place experiments.
% * <polygerinos@asu.edu> 2018-09-10T00:56:32.194Z:
% 
% >  To achieve this we evaluate the fSPL through experimentally validating the computational FEM models of the system, 
% something does not read right... revise
% Berm: Edited

\begin{figure}[t!]
% \vspace*{-0.5cm}
\centering
\includegraphics[width=0.41\textwidth]{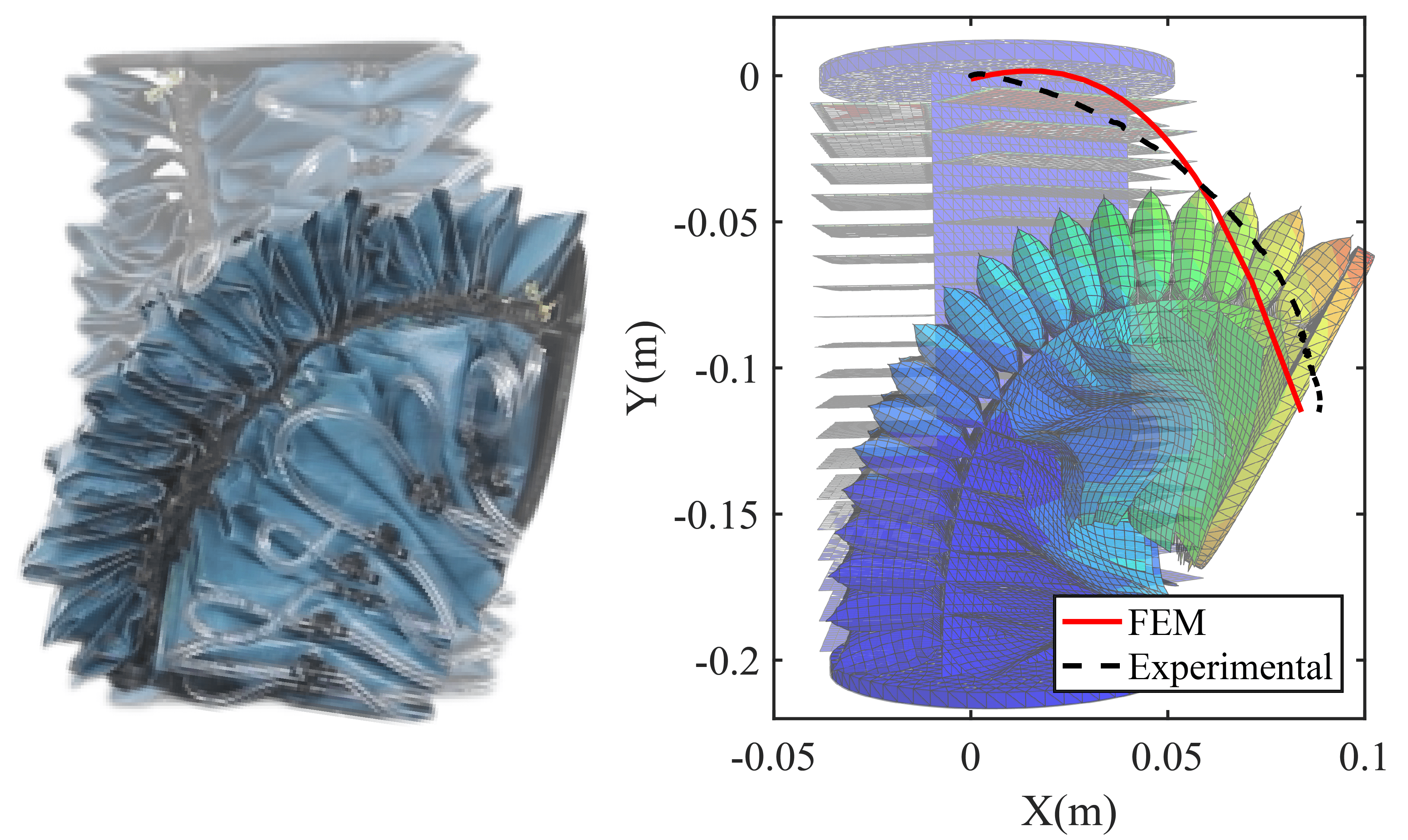}
\setlength{\belowcaptionskip}{-18pt}
\caption{Bending capability of a f3BA unit compared between FEM and experimentally validated}
\label{fig:f3CAs_bend_fem_real}

% \vspace{-1.5em}
\end{figure}

\subsection{Payload Capacity}

We have designed two experiments to investigate payload capacity of the fSPL and its components; the actuator array and f3BA. The first experiment (UTM experiment) requires to have the soft components inflate upwards against a force sensor while mounted on a universal testing machine (UTM Instron 5944, Instron Corp., High Wycombe, United Kingdom). The force output (payload) is measured at small pressure increments of 0.069$MPa$ until 0.34$MPa$ is reached. The second experiment (free-space loading experiment) requires the components to lift a set of known weights mounted at the end-effector position from a deflated state until the parallel to the ground configuration is achieved, as verified using recorded video reference. The experiment is repeated with an increased load until this configuration can no longer be achieved. The results for both experiments are summarized in Table~\ref{tab:payload_table}. Each payload experiment is repeated 3 times for the SPL and its components.
% * <polygerinos@asu.edu> 2018-09-12T00:30:42.826Z:
% 
% > This experiment is also modeled in FEM simulation and experimentally validated.
% this is a very standalone sentence. revise to try to have it flow with the rest of the text...where do the readers see this? maybe delete since you are comparing fem with experiments later in 1) 2) 3)?
% Berm: Deleted the sentence
% ^.

\subsubsection{Payload capacity of the actuator array}
\label{sec:act_array_fem_real}

The FEM and experimental data comparison for the UTM experiment are seen in Fig.~\ref{fig:single_fem_real}. Both present similar payload output trends with a root mean square error (RSME) of 9.087$N$. As seen in Table~\ref{tab:payload_table}, the single actuator array is capable of producing 177.2$N$ in the FEM simulation and 174.7$\pm$6.367$N$ experimentally, demonstrating a discrepancy error of 1.4\%.

\subsubsection{Payload capacity of the f3BA}

The load capacity when the bottom two adjacent actuator arrays of the f3BA are inflated, is shown in Fig. \ref{fig:f3CAs_load_fem_real}, with an RMSE calculated at 4.05$N$ between FEM and experiment. The maximum payload for FEM simulation and experimental validation is 56.9$N$ and 53.7$\pm$1.041$N$, respectively. Investigating the payload capacity for the f3BA in the free-space loading experiment is found capable of lifting a maximum load of 5.5$kg$, approximately 15.1x its own weight of 0.37$kg$.

% * <polygerinos@asu.edu> 2018-09-12T00:37:45.857Z:
% 
% > chambers
% still you are using the word chambers..but you havent defined anywhere
% Berm: changed to adjacent actuator arrays
% ^.

\subsubsection{Payload capacity of the fSPL}

For the UTM experiment, the payload from the fSPL FEM simulation and experimental validation is found to be 20$N$ and 14.91$\pm$0.926$N$, respectively. In this experiment, the lower adjacent chambers of the two proximal segments are pressurized up to 0.345$MPa$ while the lower adjacent chambers of the distal segment are pressurized up to 0.207$MPa$. Under the same pressurization scheme in the free-space loading experiment, the maximum payload capacity for the 1.1$kg$ fSPL is found to be 1.5$kg$, surpassing the desired payload goal of 1$kg$ set in Section~\ref{sec:func_req}.
% * <polygerinos@asu.edu> 2018-09-14T18:26:44.072Z:
% 
% > For the UTM experiment, the payload from the fSPL FEM simulation and experimental validation is found to be 20$N$ and 14.91$\pm$0.926$N$, respectively. In this experiment, the lower adjacent chambers of the two proximal segments are pressurized up to 0.345$MPa$ while the lower adjacent chambers of the distal segment are pressurized up to 0.207$MPa$. Under the same pressurization scheme in the free-space loading experiment, the maximum payload capacity for the 1.1$kg$ fSPL is found to be 1.5$kg$, surpassing the desired payload goal of 1$kg$ set in Section~\ref{sec:func_req}.
% this section is still a bit confusing ...too many things intertwined 
% 
% ^.

%in order to maintain the fSPL parallel the ground against gravity 

\begin{figure}[t!]
\centering
\includegraphics[width=0.40\textwidth]{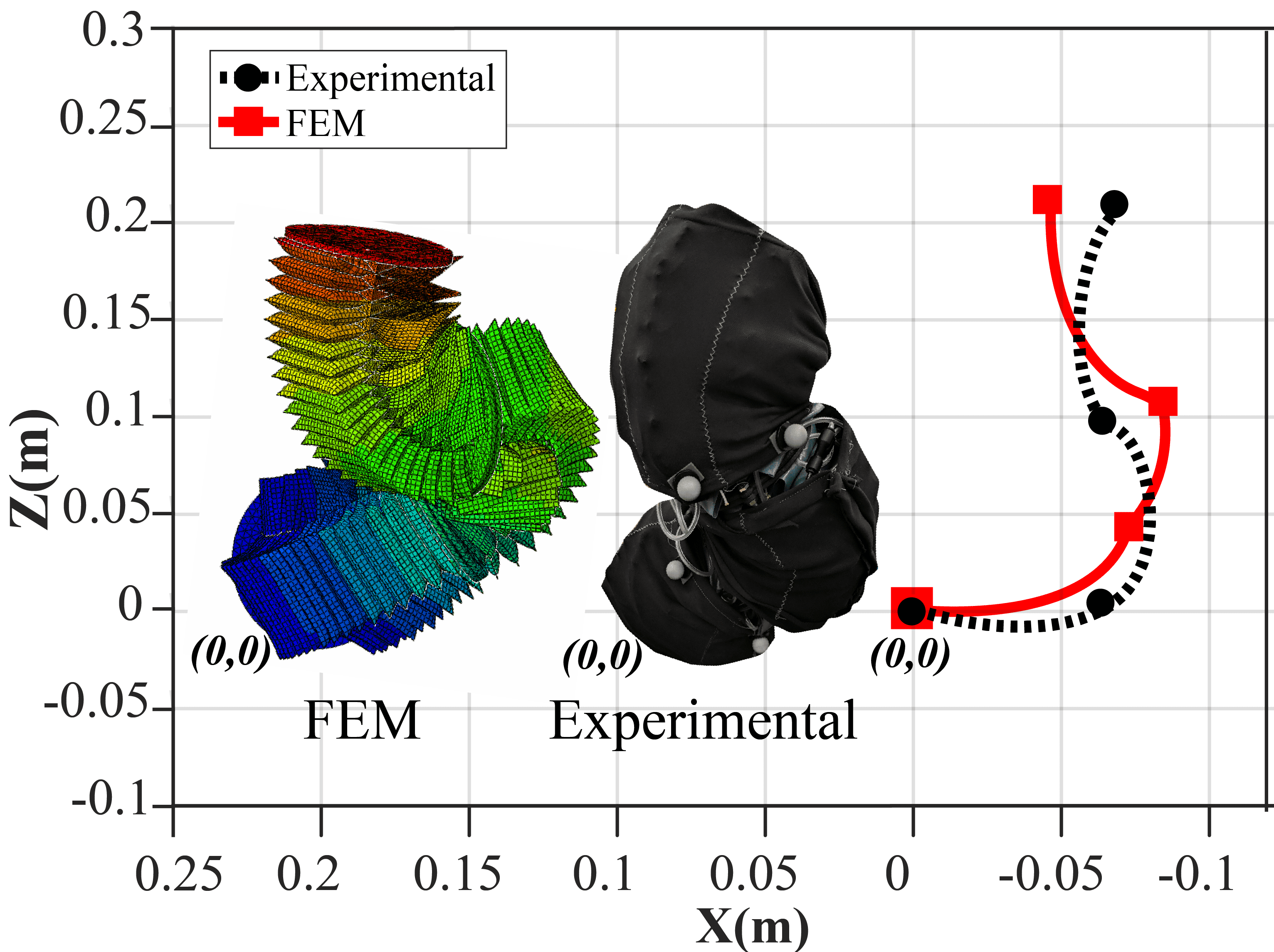}
\setlength{\belowcaptionskip}{-18pt}
\caption{Comparison of the FEM simulation vs experimentally validated pose for fSPL }
\label{fig:pose}
\end{figure}

\subsection{Motion Trajectory Tracking}

%Why are there two FEM vs Experimental sections here?
% \subsubsection{f3CBA FEM vs Experimental}
\subsubsection{f3BA Motion Trajectory}

An experiment to compare the motion trajectory of a f3BA segment of the FEM model with this of the prototype is conducted. A set of passive reflective markers are attached to the distal end of the segment while motion capture cameras (Optitrack Prime 13W, NaturalPoint Inc., Corvallis, OR) are used to track their motion when one side of the segment is inflated to 0.345$MPa$ quasi-statically, as seen in Fig. \ref{fig:f3CAs_bend_fem_real}. The end-effector position of the f3BA, in the FEM simulation and physical experiment, differ by a Euclidean distance error of 6.8$mm$. The displacement error from the initial, unpressurized f3BA segment position is 3.9\%. 

%When divided by the 17.5cm length of the f3CBA gives a displacement error of 3.9\%. %The rather close match between the trajectories of the model and experimental data shows that the FEM model could be used to predict the complex motion of a single segment of the fSPL.
% * <polygerinos@asu.edu> 2018-09-12T01:01:54.216Z:
% 
% > 3CA
% 3CA?????
% Berm: 3fCBA
% ^.
% * <polygerinos@asu.edu> 2018-09-10T01:15:31.986Z:
% 
% > uclidean distance error of 6.8$mm$
% what does that mean in terms of % error?
% Berm: Added Percentage Error
% ^.
% * <polygerinos@asu.edu> 2018-09-10T01:14:14.557Z:
% 
% > in the center of the distal enter 
% what are you trying to say here? revise
% Berm: Revised
% ^.

\begin{figure}[b!]

\vspace*{-0.5cm}
\centering
\includegraphics[width=0.43\textwidth]{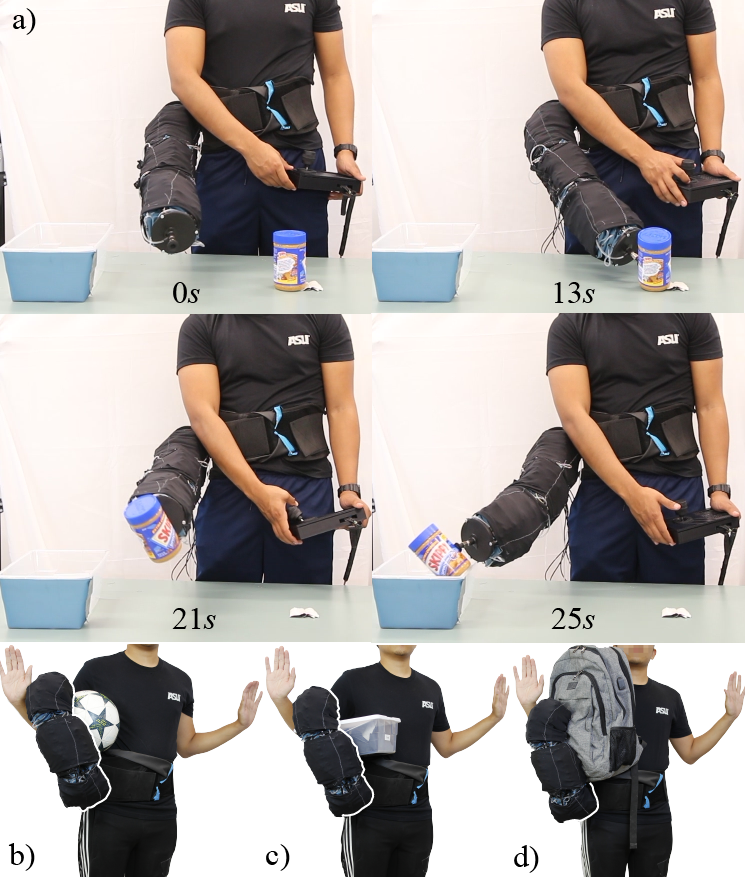}
\caption{a) Pick-and-place test using suction end-effector with fSPL. Wrap around grasping method for a b) ball c) box d) bag.}
% * <polygerinos@asu.edu> 2018-09-14T20:48:42.639Z:
% 
% > The different methods the fSPL can carry payload, either using its end-effector or whole body to wrap around objects.}
% add letters to guide the reader through the sequence of images
% 
% ^.
\label{fig:pick_place_whole_body}
% \vspace{-1.5em}
\end{figure}
\subsubsection{fSPL Motion Trajectory}
To study the FEM model's ability to predict the non-linear motion behavior produced by the fSPL, a highly complex pose of the fSPL is generated arbitrarily and recorded using the motion capture system, as shown in Fig.~\ref{fig:pose}. Markers are added at the proximal and distal end of every segment to create virtual rigid bodies. The pose is achieved by inflating the alternating bottom actuator arrays of each segment to 0.207$MPa$ respectively in FEM. The experimental pose is achieved by inflating the same actuator sequence by 0.207$MPa$, 0.241$MPa$, and 0.172$MPa$ subsequently. There is a slight variation in the inflation of the second and third segment by $\pm$0.0345$MPa$ to compensate for the differences in FEM and reality, like manufacturing inconsistencies.The experimental Euclidean displacement error is calculated at 3.96$cm$ with FEM model estimated pose and had a 7.55\% displacement error. 

% * <polygerinos@asu.edu> 2018-09-12T01:06:56.164Z:
%  and observed buckling not predicted by FEM
% >  bladder 
% bladder again...change
% Berm: DONE
% ^.
% * <polygerinos@asu.edu> 2018-09-12T01:03:53.315Z:
% 
% > We determine the FEM model to predict the continuum complex non-linear motion produced by the entire fSPL, a single complex pose is generated and recorded using the motion capture system. 
% revise...this sentence does not make sense
% Revised: Berm
% ^.
% * <polygerinos@asu.edu> 2018-09-10T01:16:39.488Z:
% 
% > We determine the FEM model to predict the continuum complex non-linear motion produced by the entire fSPL, a single complex pose is generated and recorded using the motion capture system. 
% revise...not clear 
% Berm: revised
% ^.

% \subsection{Payload Capacity of the fSPL}

% \begin{figure}[t!]
% \centering
% \includegraphics[width=0.45\textwidth]{Figures/payload_all_v2}
% \caption{Payload Setup for single bladder array, f3CBA, and fSPL}
% \label{fig:payload_all}
% \vspace{-1.5em}
% \end{figure}

\subsection{Pick-and-Place User Experiments}

To test the motion performance of the fSPL when used in real-life situations, along with user controllability, a pick-and-place experiment is performed with a number of human participants ($n$=3). For this experiment, a vacuum suction cup is used at the fSPL's end-effector. The vacuum suction cup has a theoretical payload of 0.86$kg$ and is connected to a vacuum pump with depressurization rate of 1.42x10$^{-3}m^3/s$. The experiment is set up for the user to operate the fSPL using a joystick controller previously introduced \cite{nguyen2018}, to pick a jar with peanut butter (0.3$kg$) on one end of a table and move it across and inside a target box (0.33x0.16x0.12$m$) that is placed 0.45$m$ away, as seen in Fig. \ref{fig:pick_place_whole_body}. Three first-time users are timed while performing the task five times each.  %User 1 improved their time from 53.28$s$ to 24.70$s$, user 2 improved from 45.34$s$ to 20.79$s$, and user 3 improved their time from 27.69$s$ to 17$s$. 
The timed results are shown in table \ref{tab:pickplace_table} demonstrating the gradual adoption and improvement in controlling a soft, external limb. A variety of other daily livings objects are also successfully picked and placed, including a hair spray canister  (0.43$kg$) and a water bottle (0.88$kg$).
% * <polygerinos@asu.edu> 2018-09-12T01:11:32.802Z:
% 
% > (0.1$kg$)
% is this correct? seems low
% Berm: DONE
% ^.
% * <polygerinos@asu.edu> 2018-09-10T01:26:13.104Z:
% 
% > using a joystick controller to
% reviewers will ask about the controller...either reference past work or add a brief explanation here
% Berm: Added citation
% ^.
% * <polygerinos@asu.edu> 2018-09-10T01:25:29.851Z:
% 
% > eir time from 53.28$s$ to 24.70$s$, user 2 improved from 45.34$s$ to 20.79$s$, and user 3 improved their time from 27.69$s$ to 17$s$. Thus preliminary demonstrating the gradual adoption and improvement in controlling a soft, external limb. A variety of other daily livings objects are also successfully picked and placed, including a canister of hair spray (0.43$kg$) and a water bottle 0.88$kg$.
%
%
% you need a table that shows all three objects and subsequent times performed in each trial for all users
% We just did one object, the peanut butter cup. 
% ^.
% * <polygerinos@asu.edu> 2018-09-10T01:22:48.354Z:
% 
% >  for multiple users.
% BERM: how many? 3 users
% 
% ^.

\begin{table}[t!]
\caption{Pick-and-Place Time Duration} 
\label{tab:pickplace_table}
	\begin{tabularx}{0.48\textwidth}{c|c|c|c|c|c}   \toprule\toprule
    \centering
    \small
    \setlength\tabcolsep{11pt}
	\textbf{\emph{Users}} & \textbf{\emph{Attempt  }} & \textbf{\emph{Attempt }} & \textbf{\emph{Attempt }} & \textbf{\emph{Attempt  }} & \textbf{\emph{Attempt }}\\ 
  	
   						 & \textbf{\emph{1 }} 		& 	\textbf{\emph{2}} 			& \textbf{\emph{3 }} & 	\textbf{\emph{4 }} & \textbf{\emph{ 5 }}\\   
    
    \midrule
	1	&	53.28$s$	& 	37.28$s$ 	& 	27.17$s$ 	&	25.16$s$ 	&	24.70$s$ 	\\
   	2  	&	45.34$s$   &  	33.88$s$	& 	22.33$s$ 	&	21.43$s$ 	& 	20.79$s$ 	\\
    3 	&	27.68$s$	&	21.89$s$	&	19.60$s$	&	17.00$s$	&	13.12$s$	\\
    \bottomrule 
    \hline
	\end{tabularx}
    \vspace{-1.5em}
\end{table}

% \subsubsection{Fold and Unfold}
% Our preliminary designs show that such compliant segment is able to linearly collapse two times its own length when squeezed, but also linearly deploy and perform multi-degree-of-freedom bending motions when pressurized (Figure 14).

% \begin{figure}[t!]
% \centering
% \includegraphics[width=0.30\textwidth]{Figures/unfold_fold}
% \caption{ Side view of soft fabric segment comprised of three bending
% soft fabric bundles in a triangle formation, able to collapse four 
% times its length and bend in multi-DoF when deployed and ressurized.}
% \label{fig:unfold_fold}
% \vspace{-1.5em}
% \end{figure}

\subsection{Whole-Body Continuum Grasping}

Similar to an elephant using its trunk, the fSPL is also capable of manipulating objects, larger than the size of the end-effector, by wrapping itself around them. In order to identify the effective whole-body grasping tactic, a trial and error process was used. The most efficient grasping method to carry heavier and larger objects is shown in Fig.~\ref{fig:pick_place_whole_body}b, where the majority of the load is supported by the proximal segment, while creating a tight wrap around the object against the body of the user that is standing upright. To demonstrate this feature, the fSPL is attached to a belt worn by a user and inflated to wrap around various objects. The objects tested include a soccer ball (0.45$kg$), a box (1.75$kg$), and a backpack (initially 1.61$kg$), as seen in Fig.~\ref{fig:pick_place_whole_body}. The maximum payload capacity using this method is tested by gradually adding weights into the backpack resulting in a load of 11.13$kg$ (10.1x its own body weight) before the fSPL loses support.
% * <polygerinos@asu.edu> 2018-09-14T18:39:15.955Z:
% 
% > The maximum payload capacity using this method is tested by gradually adding weights into the backpack resulting in a load of 11.13$kg$ (10.1x its own body weight) before the fSPL loses support.
% a brief comment is needed here on how these heavy loads are achieved and that size of object and grasping configurations need to be explored for different objects to ensure adequate grasping...
% 
% ^.
% * <polygerinos@asu.edu> 2018-09-14T18:37:03.296Z:
% 
% > This method of grasping allows the fSPL to carry heavier and larger objects, when compared to its end-effector, by supporting the majority of the load at its proximal segment, while creating a tight wrap around the object against the body of the user that is standing upright, using its continuum body.
% very long sentence revise
% 
% ^.

\subsection{Workspace}

\begin{figure}[t!]
% * <polygerinos@asu.edu> 2018-09-15T03:38:56.623Z:
% 
% > \begin{figure}[t!]
% > \vspace*{-0.5cm}
% > \centering
% > \includegraphics[width=0.43\textwidth]{Figures/Workspace_Fabric_ARM}
% > \setlength{\belowcaptionskip}{-18pt}
% > \caption{The workspace of the SPL in the coronal plane.}
% 
% 
% in the text somewhere you are mentioning that the effective workspace  is 60 cm by 60cm ...this isnt the case when someone looks at the graph...make sure you are consistent. 
% 
% ^.
% \vspace*{-0.5cm}
\centering
\includegraphics[width=0.41\textwidth]{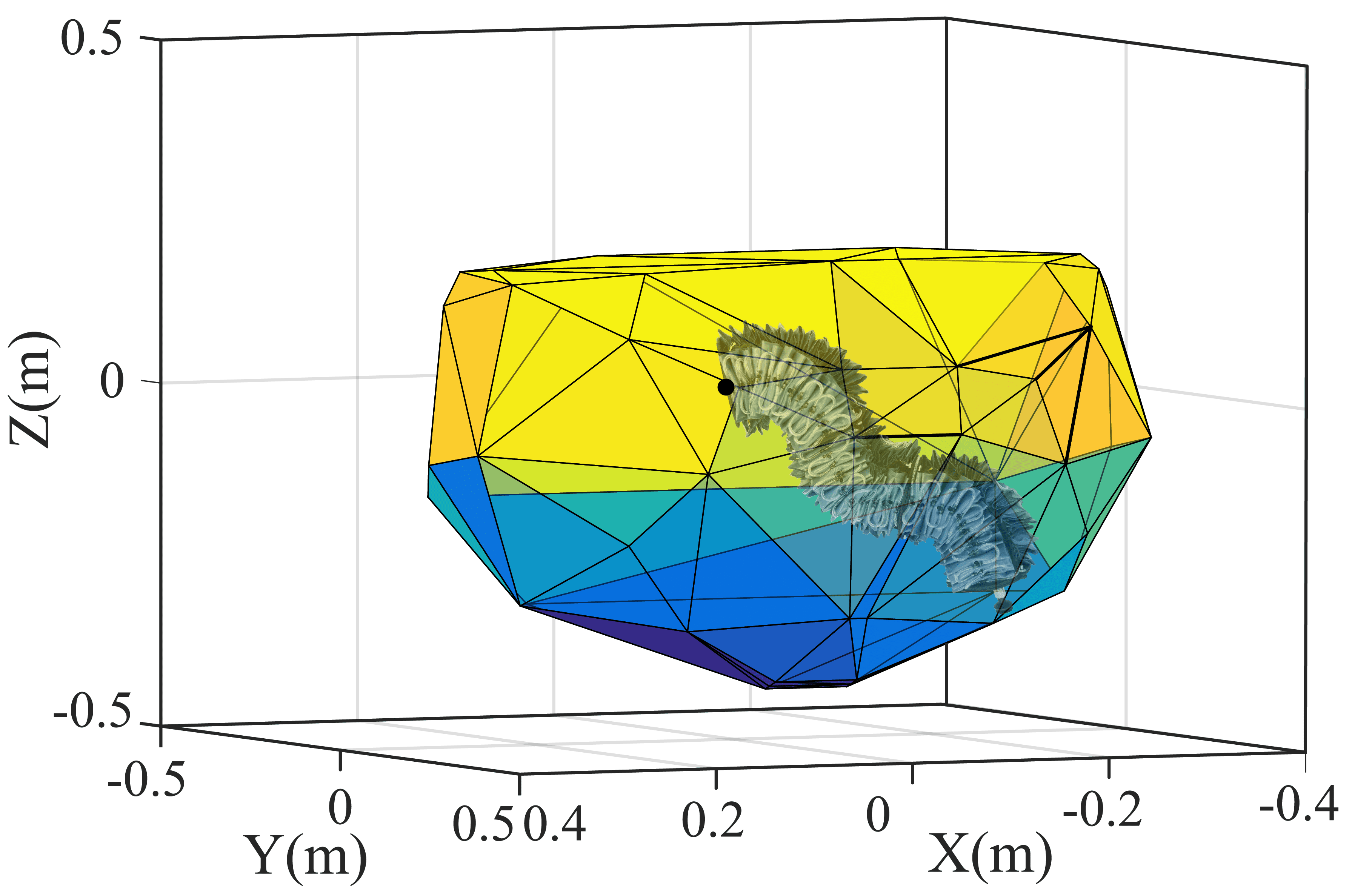}
\setlength{\belowcaptionskip}{-18pt}
\caption{The workspace of the fSPL in the coronal plane.}
\label{fig:workspace_fabric_arm}
%  \vspace{-0.5em}
\end{figure}

To evaluate the effective workspace, the fSPL is mounted parallel to the ground. Two sets of reflective markers are placed at the distal and proximal ends of the fSPL to create virtual rigid bodies. The position of the markers is recorded using a motion capture system. The individual actuator arrays of the limb are inflated in various configurations using a maximum pressure of 0.34$MPa$ to cover their entire workspace. The workspace of the fSPL is generated from the collected end-effector positions shown in Fig \ref{fig:workspace_fabric_arm}. The fSPL is shown capable of reaching any point inside this workspace, where the shaded colors show the change in height in the Z-axis. Resulting in an effective volume of 0.123$m^{3}$, with a maximum vertical range of 0.63$m$ and maximum horizontal range of 0.695$m$. This workspace provides enough vertical and horizontal range to support a user in ADL tasks in the coronal plane, based on how it is mounted. This orientation can be adjusted based on the task and user preference to optimize the effective workspace.

\section{Conclusion and Future Work}

In this paper, the design, development, and evaluation of a novel, user-mounted, fabric Soft Poly-Limb (fSPL) was introduced utilizing high-strength inflatable fabrics. The highly-articulated, soft fSPL was designed towards supporting and assisting individuals to perform ADL. The fSPL's geometrical parameters were optimized and preliminary experimentally validated using quasi-static computational FEM models, providing design guidelines for the community to design fSPLs based on the desired payload capacity and articulation performance.

Each fSPL segment was capable of effectively lifting 15.1x its own weight at 5.5$kg$ and the entire fSPL demonstrated the capacity to carry up to 1.5$kg$ of load (1.5x its total weight) while able to maneuver in space. By employing the whole body grasping methodology, the fSPL was shown able to wrap around and carry loads of up to 11.13$kg$, over 10.1 times it’s own body weight. Furthermore, the fSPL demonstrated its complex motion abilities by operating when worn by users while offering a large workspace and safely interacting with the environment. 
% * <polygerinos@asu.edu> 2018-09-14T18:45:53.368Z:
% 
% >  11.13$kg$, o
% double check your numbers that they match
% 
% ^.

Future work will introduce the exploration of multi-mounting positions and multiple interacting fSPLs on users to expand the workspace and capabilities of controlling multi-external limbs. Further exploration of various user-intent detection methods will also be explored to control the fSPL, hands-free. Implementation of on-board actuation and power supply systems that are capable of controlling such a complicated system are being explored. Future efforts are being made to control the SPL using distributed sensing and control methods \cite{wenlong2017} with the exploration of soft-distributed and embedded sensing technologies as well. 
% * <polygerinos@asu.edu> 2018-09-10T01:37:21.810Z:
% 
% > Implementation of a decentralized and distributed on-board actuation and power supply systems that are capable of controlling such a complicated system is being explored. Future efforts are being made to control the SPL using distributed sensing and control methods \cite{Wenlong2018} with the exploration of soft-distributed and embedded sensing technologies as well. 
% these two statements sound a bit repetitive with one another..please revise

% Revised
% ^.

% In total, we believe that this work provides the readers with the capability of designing highly lightweight and soft robotic limbs capable of being easily stored and deployed on-site capable of support the versatile amount of activities of daily living.

%%%%%%%%%%%%%%%%%%%%%%%%%%%%%%%%%%%%%%%%%%%%%%%%%%%%%%%%%%%%%%%%%%%%%%%%%%%%%%%%

\section*{ACKNOWLEDGMENTS} 
This work was supported in part by the National Science Foundation under Grant CMMI-1800940.

%%%%%%%%%%%%%%%%%%%%%%%%%%%%%%%%%%%%%%%%%%%%%%%%%%%%%%%%%%%%%%%%%%%%%%%%%%%%%%%%

%%%%%%%%%%%%%%%%%%%%%%%%%%%%%%%%%%%%%%%%%%%%%%%%%%%%%%%%%%%%%%%%%%%%%%%%%%%%%%%%
%%%%%%%%%%%%%%%%%%%%%%%%%%%%%%%%%%%%%%%%%%%%%%%%%%%%%%%%%%%%%%%%%%%%%%%%%%%%%%%%

% ********************************** Bibliography ******************************
\bibliographystyle{unsrt}
%\bibliographystyle{plain}
%\bibliographystyle{plainnat} % use this to have URLs listed in References
%\cleardoublepage
\bibliography{ICRA2019} % Path to your References.bib file

\begin{thebibliography}{10}

\bibitem{gopura2011}
R~A R~C Gopura, Kazuo Kiguchi, and D~S~V Bandara.
\newblock {A brief review on upper extremity robotic exoskeleton systems}.
\newblock In {\em 2011 6th International Conference on Industrial and
  Information Systems}, volume 8502, pages 346--351, aug 2011.

\bibitem{bogue2009}
Robert Bogue and Robert Bogue.
\newblock {Exoskeletons and robotic prosthetics: a review of recent
  developments}.
\newblock {\em Industrial Robot: the international journal of robotics research
  and application}, 36(5):421--427, 2009.

\bibitem{kurek2017}
Daniel~A Kurek and H~Harry Asada.
\newblock {The MantisBot: Design and Impedance Control of Supernumerary Robotic
  Limbs for Near-Ground Work}.
\newblock In {\em 2017 IEEE International Conference on Robotics and Automation
  (ICRA)}, pages 5942--5947, may 2017.

\bibitem{parietti2015}
Federico Parietti, Kameron~C. Chan, Banks Hunter, and H.~Harry Asada.
\newblock {Design and control of Supernumerary Robotic Limbs for balance
  augmentation}.
\newblock In {\em Proceedings - IEEE International Conference on Robotics and
  Automation}, volume 2015-June, pages 5010--5017, may 2015.

\bibitem{parietti2014}
Federico Parietti and H.~Harry Asada.
\newblock {Supernumerary Robotic Limbs for aircraft fuselage assembly: Body
  stabilization and guidance by bracing}.
\newblock In {\em 2014 IEEE International Conference on Robotics and Automation
  (ICRA)}, pages 1176--1183, may 2014.

\bibitem{saraiji2018}
M~H D~Yamen Saraiji, Tomoya Sasaki, Reo Matsumura, Kouta Minamizawa, and
  Masahiko Inami.
\newblock {Fusion: Full Body Surrogacy for Collaborative Communication}.
\newblock In {\em ACM SIGGRAPH 2018 Emerging Technologies}, SIGGRAPH '18, pages
  7:1----7:2, New York, NY, USA, 2018. ACM.

\bibitem{vatsal2017}
Vighnesh Vatsal and Guy Hoffman.
\newblock {Wearing Your Arm on Your Sleeve : Studying Usage Contexts for a
  Wearable Robotic Forearm}.
\newblock In {\em 2017 26th IEEE International Symposium on Robot and Human
  Interactive Communication (RO-MAN)}, pages 974--980, aug 2017.

\bibitem{hussain2017b}
Irfan Hussain, Gionata Salvietti, Giovanni Spagnoletti, Monica Malvezzi, David
  Cioncoloni, Simone Rossi, and Domenico Prattichizzo.
\newblock {A soft supernumerary robotic finger and mobile arm support for
  grasping compensation and hemiparetic upper limb rehabilitation}.
\newblock {\em Robotics and Autonomous Systems}, 93:1--12, 2017.

\bibitem{wu2015}
Faye~Y. Wu and H.~Harry Asada.
\newblock {'Hold-and-manipulate' with a single hand being assisted by wearable
  extra fingers}.
\newblock In {\em Proceedings - IEEE International Conference on Robotics and
  Automation}, volume 2015-June, pages 6205--6212, 2015.

\bibitem{tiziani2017}
Lucas Tiziani, Alexander Hart, Thomas Cahoon, Faye Wu, H~Harry Asada, and
  Frank~L Hammond.
\newblock {Empirical characterization of modular variable stiffness inflatable
  structures for supernumerary grasp-assist devices}.
\newblock {\em The International Journal of Robotics Research},
  36(13-14):1391--1413, 2017.

\bibitem{delAma2012}
Antonio~J. Del-Ama, Aikaterini~D. Koutsou, Juan~C. Moreno, Ana De-los Reyes,
  Ngel Gil-Agudo, and Jos~L. Pons.
\newblock {Review of hybrid exoskeletons to restore gait following spinal cord
  injury}.
\newblock {\em The Journal of Rehabilitation Research and Development},
  49(4):497, 2012.

\bibitem{mcMahan2005}
William McMahan, Bryan~A. Jones, and Ian~D. Walker.
\newblock {Design and implementation of a multi-section continuum robot:
  Air-octor}.
\newblock In {\em 2005 IEEE/RSJ International Conference on Intelligent Robots
  and Systems, IROS}, number January, pages 3345--3352, aug 2005.

\bibitem{calisti2011}
M.~Calisti, M.~Giorelli, G.~Levy, B.~Mazzolai, B.~Hochner, C.~Laschi, and
  P.~Dario.
\newblock {An octopus-bioinspired solution to movement and manipulation for
  soft robots}.
\newblock {\em Bioinspiration {\&} Biomimetics}, 6(3):36002, 2011.

\bibitem{walker2005}
Ian~D. Walker, Darren~M. Dawson, Tamar Flash, Frank~W. Grasso, Roger~T. Hanlon,
  Binyamin Hochner, William~M. Kier, Christopher~C. Pagano, Christopher~D.
  Rahn, and Qiming~M. Zhang.
\newblock {Continuum robot arms inspired by cephalopods}.
\newblock {\em SPIE Conference on Unmanned Ground Vehicle Technology},
  5804:303--314, 2005.

\bibitem{godage2016}
Isuru~S. Godage, Gustavo~A. Medrano-Cerda, David~T. Branson, Emanuele
  Guglielmino, and Darwin~G. Caldwell.
\newblock {Dynamics for variable length multisection continuum arms}.
\newblock {\em The International Journal of Robotics Research}, 35(6):695--722,
  2016.

\bibitem{yasmin2017}
Yasmin Ansari, Mariangela Manti, Egidio Falotico, Yoan Mollard, Matteo
  Cianchetti, and Cecilia Laschi.
\newblock {Towards the development of a soft manipulator as an assistive robot
  for personal care of elderly people}.
\newblock {\em International Journal of Advanced Robotic Systems},
  14(2):1729881416687132, 2017.

\bibitem{giannaccini2017}
Maria~Elena Giannaccini, Chaoqun Xiang, Adham Atyabi, Theo Theodoridis, Samia
  Nefti-Meziani, Steve Davis, Giannaccini~Maria Elena, Xiang Chaoqun, Atyabi
  Adham, Theodoridis Theo, Nefti-Meziani Samia, and Davis Steve.
\newblock {Novel Design of a Soft Lightweight Pneumatic Continuum Robot Arm
  with Decoupled Variable Stiffness and Positioning}.
\newblock {\em Soft Robotics}, 00(00):soro.2016.0066, 2017.

\bibitem{cianchetti2013}
Matteo Cianchetti, Tommaso Ranzani, Giada Gerboni, Iris~De Falco, Cecilia
  Laschi, Senior Member, and Arianna Menciassi.
\newblock {STIFF-FLOP surgical manipulator: Mechanical design and experimental
  characterization of the single module}.
\newblock In {\em 2013 IEEE/RSJ International Conference on Intelligent Robots
  and Systems}, pages 3576--3581, nov 2013.

\bibitem{marchese2015}
Andrew~D. Marchese and Daniela Rus.
\newblock {Design, kinematics, and control of a soft spatial fluidic elastomer
  manipulator}.
\newblock {\em The International Journal of Robotics Research},
  35(7):0278364915587925--, 2015.

\bibitem{robertson2017}
Matthew~A Robertson and Jamie Paik.
\newblock {New soft robots really suck: Vacuum-powered systems empower diverse
  capabilities}.
\newblock {\em Science Robotics}, 2(9):1--12, 2017.

\bibitem{gong2018}
Zheyuan Gong, Jiahui Cheng, Xingyu Chen, Wenguang Sun, Xi~Fang, and Kainan Hu.
\newblock {A Bio-inspired Soft Robotic Arm : Kinematic Modeling and
  Hydrodynamic Experiments}.
\newblock {\em Journal of Bionic Engineering}, 15:204--219, 2018.

\bibitem{santoso2017}
Junius Santoso, Erik~H Skorina, Ming Luo, Ruibo Yan, and Cagdas~D Onal.
\newblock {Design and analysis of an origami continuum manipulation module with
  torsional strength}.
\newblock In {\em IEEE International Conference on Intelligent Robots and
  Systems}, volume 2017-Septe, pages 2098--2104, 2017.

\bibitem{sanan2013}
Siddharth Sanan.
\newblock {\em {Soft Inflatable Robots for Safe Physical Human Interaction}}.
\newblock PhD thesis, Carnegie Mellon University, Pittsburgh, PA, aug 2013.

\bibitem{hawkes2017}
Elliot~W Hawkes, Laura~H Blumenschein, Joseph~D Greer, and Allison~M Okamura.
\newblock {A soft robot that navigates its environment through growth}.
\newblock {\em Science Robotics}, 2(8):1--8, 2017.

\bibitem{ohta2017}
Ohta Preston, Valle Luis, King Jonathan, Low Kevin, Yi~Jaehyun,
  Atkeson~Christopher G., Park Yong-Lae, Preston Ohta, Luis Valle, Jonathan
  King, Kevin Low, Jaehyun Yi, Christopher~G. Atkeson, and Yong-Lae Park.
\newblock {Design of a Lightweight Soft Robotic Arm Using Pneumatic Artificial
  Muscles and Inflatable Sleeves}.
\newblock {\em Soft Robotics}, 0(0):null, 2017.

\bibitem{best2016}
Charles~M. Best, Morgan~T Gillespie, Phillip Hyatt, Levi Rupert, Vallan
  Sherrod, and Marc~D Killpack.
\newblock {A New Soft Robot Control Method: Using Model Predictive Control for
  a Pneumatically Actuated Humanoid}.
\newblock {\em IEEE Robotics {\&} Automation Magazine}, 23(3):75--84, 2016.

\bibitem{liang2018}
Xinquan Liang, Haris Cheong, Yi~Sun, Jin Guo, Chee~Kong Chui, and C~Yeow.
\newblock {Design , Characterization and Implementation of a Two - DOF Fabric -
  based Soft Robotic Arm}.
\newblock {\em IEEE Robotics and Automation Letters}, 3766(c):1--8, jul 2018.

\bibitem{takeichi2017}
Masashi Takeichi, Koichi Suzumori, Gen Endo, and Hiroyuki Nabae.
\newblock {Development of Giacometti Arm With Balloon Body}.
\newblock {\em IEEE Robotics and Automation Letters}, 2(2):2710--2716, 2017.

\bibitem{kim2018}
Hye~Jong Kim, Akihiro Kawamura, Yasutaka Nishioka, and Sadao Kawamura.
\newblock {Mechanical design and control of inflatable robotic arms for high
  positioning accuracy}.
\newblock {\em Advanced Robotics}, 32(2):89--104, 2018.

\bibitem{nguyen2018}
Pham~Huy Nguyen, Curtis Sparks, Sai~Gautham Nuthi, Nicholas~M Vale, and
  Panagiotis Polygerinos.
\newblock {Soft Poly-Limbs: Toward a New Paradigm of Mobile Manipulation for
  Daily Living Tasks}.
\newblock {\em Soft Robotics}, 00(00):soro.2018.0065, 2018.

\bibitem{carly2018}
Carly~M. Thalman, Quoc~P. Lam, Pham~H. Nguyen, Saivimal Sridar, and Panagiotis
  Polygerinos.
\newblock {A Novel Soft Elbow Exosuit to Supplement Bicep Lifting Capacity}.
\newblock In {\em 2018 IEEE/RSJ International Conference on Intelligent Robots
  and Systems, IROS}, 2018.
\newblock [Accepted].

\bibitem{plagenhoef1983}
Stanley Plagenhoef, F.~Gaynor Evans, and Thomas Abdelnour.
\newblock {Anatomical Data for Analyzing Human Motion}.
\newblock {\em Research Quarterly for Exercise and Sport}, 54(2):169--178,
  1983.

\bibitem{wenlong2017}
W.~Zhang and P.~Polygerinos.
\newblock Distributed planning of multi-segment soft robotic arms.
\newblock In {\em 2018 Annual American Control Conference (ACC)}, pages
  2096--2101, June 2018.

\end{thebibliography}

\end{document}